\documentclass[11pt,a4paper]{article}
\usepackage[hyperref]{emnlp2020}
\usepackage{times}
\usepackage{latexsym}

\usepackage{microtype}
\usepackage{graphicx}
\usepackage{booktabs} % for professional tables
\usepackage{amsmath}
\usepackage{subcaption}

\newcommand{\closureex}[1]{{#1}}

\aclfinalcopy

\title{CLOSURE: Assessing Systematic Generalization of CLEVR Models}

\author{Dzmitry Bahdanau\thanks{~~Work mostly done during PhD studies at Universit\'e de Montr\'eal and Mila.} \textsuperscript{12}
  Harm de Vries\textsuperscript{1}
  Timothy J. O’Donnell\textsuperscript{4}
  Shikhar Murty\textsuperscript{5}
  \\
  \textbf{Philippe Beaudoin\textsuperscript{1}}
  \textbf{Yoshua Bengio\textsuperscript{236}}
  \textbf{Aaron Courville\textsuperscript{236}}
  \\
  \textsuperscript{1}Element AI
  ~~~~~~\textsuperscript{2}Quebec Artificial Intelligence Institute (Mila) \\
  ~~~\textsuperscript{3}Universit\'e de Montr\'eal
  ~~~\textsuperscript{4}McGill University
  ~~~\textsuperscript{5}Stanford University  
  ~~~\textsuperscript{6}CIFAR Fellow\\
}

\begin{document}

\maketitle

\begin{abstract}
The CLEVR dataset of natural-looking questions about 3D-rendered scenes has recently received much attention from the research community. A number of models have been proposed for this task, many of which achieved very high accuracies of around 97-99\%. In this work, we study how systematic the generalization of such models is, that is to which extent they are capable of handling novel combinations of known linguistic constructs. To this end, we test models' understanding of referring expressions based on matching object properties (such as e.g. ``another cube that is the same size as the brown cube'') in novel contexts. Our experiments on the thereby constructed CLOSURE benchmark show that state-of-the-art models often do not exhibit systematicity after being trained on CLEVR. Surprisingly, we find that an explicitly compositional Neural Module Network model also generalizes badly on CLOSURE, even when it has access to the ground-truth programs at test time.  We improve the NMN's systematic generalization by developing a novel Vector-NMN module architecture with vector-valued inputs and outputs. Lastly, we investigate how much few-shot transfer learning can help models that are pretrained on CLEVR to adapt to CLOSURE. Our few-shot learning experiments contrast the adaptation behavior of the models with intermediate discrete programs with that of the end-to-end continuous models.
\end{abstract}

\begin{figure}[t]
    \begin{center}
        \includegraphics[width=1.0\linewidth]{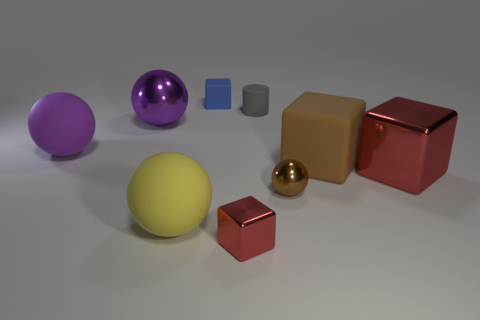}
    \end{center}
    \small
    \textbf{Q1 (CLEVR):} There is \textcolor{red}{another cube that is the same size as the brown cube}; what is its color?  \\
    \textbf{Q2 (CLEVR):} There is a thing that is in front of the yellow thing; does it have the same color as cylinder? \\
    \textbf{Q3 (CLOSURE):}  There is \textcolor{red}{another rubber object that is the same size as the gray cylinder}; does it have the same color as the tiny shiny block?
    \caption{CLEVR questions (Q1 and Q2) require complex multi-step reasoning about the contents of 3D-rendered images. We construct CLOSURE questions (Q3) by using the referring expressions that rely on matching object properties (e.g. the red fragment in Q1) in novel contexts, such as e.g. comparison questions with two referring expressions (Q2).}
    \label{fig:closure}
    \vspace*{-4mm}
\end{figure}

\section{Introduction}
The ability to communicate in natural language and ground it effectively into our rich unstructured 3D reality is a crucial skill that we expect from artificial agents of the future. A popular task to benchmark progress towards this goal is \emph{Visual Question Answering} (VQA), in which one must give a (typically short) answer to a question about the content of an image. 
The release of the relatively large VQA~1.0 dataset by \citet{antol_vqa:_2015} ignited the interest for the VQA setup, but researchers soon found that the biases of natural data (such as the heavily skewed answer distribution for certain question types) make it hard to interpret the VQA~1.0 results~\citep{agrawal_analyzing_2016}. To complement biased natural data, \citet{johnson_clevr:_2016} constructed the CLEVR dataset of complex synthetic questions about 3D-rendered scenes to be free of such biases (see Q1 and Q2 in Figure \ref{fig:closure} for examples of CLEVR questions). 
The CLEVR dataset has spurred VQA modeling research, and many models were designed for it and showcased using it \citep{santoro_simple_2017,perez_learning_2017,johnson_inferring_2017,hudson_compositional_2018,mascharka_transparency_2018}
%vedantam_probabilistic_2019}.

The high complexity and diversity of CLEVR questions and the reported 97-99\% accuracies may lead to the impression that these high-performing models are capable of answering any possible question that uses the same linguistic constructs as in CLEVR. Such an intuitive expectation corresponds to the concept of \textit{systematicity} \citep{fodor_connectionism_1988}, which characterizes the ability of humans to interpret arbitrary combinations of known primitives. One can argue that systematicity is also a highly desirable property for AI systems. For example, suppose you refer to an object by relating its appearance to another object, as  in ``the object that is the same size as the brown cube''. If a CLEVR-trained model understands such a referring expression, it is likely that you will expect this model to understand it in other, more complex contexts. This includes cases in which such an expression is embedded in a more complex referring expression, e.g. ``the cylinder to the left of the object that is the same size as the brown cube'', or is logically combined with another expression, e.g. ``either cubes or objects that are the same size as the brown cube''.  For learning-based systems, unless the training distribution uniformly covers all sensible  compositions  of  interest  (which  is  nearly impossible  to achieve for natural data),
systematicity requires a particular kind of out-of-distribution generalization, whereby the  test  distribution  is  different  from  the  training  one  but follows the same rules of semantic and syntactic composition. Such \textit{systematic generalization} of modern neural models has been recently studied in the context of artificial sequence transduction and VQA tasks \citep{lake_generalization_2018,bahdanau_systematic_2019}, the latter done in a setup that is much simpler and less diverse than CLEVR.

\begin{figure}
    \centering
    \includegraphics[scale=0.3]{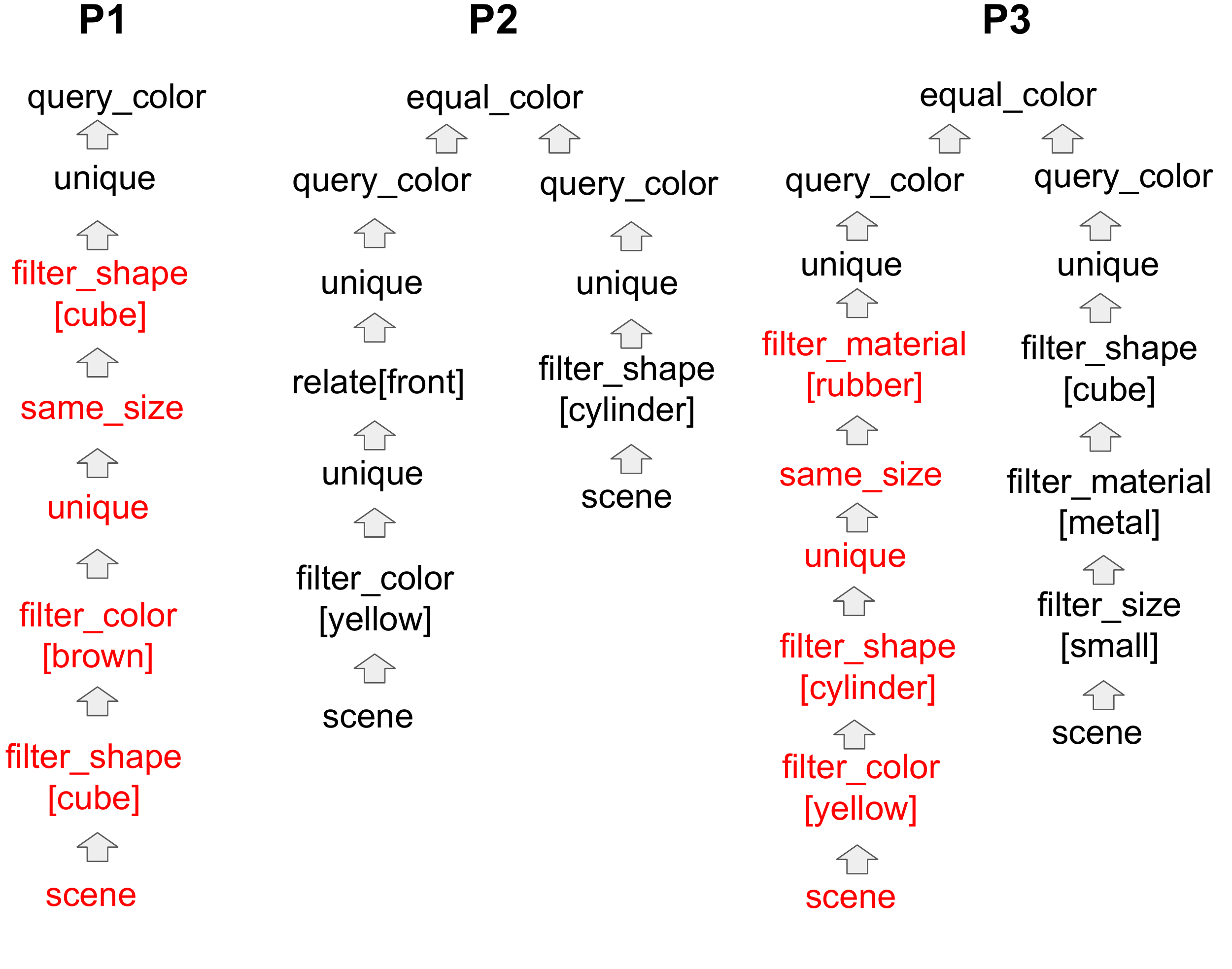}
    \vspace{-0.5cm}
    \caption{Programs P1, P2, P3 that define the ground-truth meaning for the questions Q1, Q2 and Q3 in Figure \ref{fig:closure}. The fragments in red correspond to the matching REs in the respective questions.}
    \label{fig:programs}
    \vspace*{-4mm}
\end{figure}

In this work, we perform a case-study of how systematic CLEVR-trained models are in their generalization capabilities. In doing so, we seek to provide important context to the near-perfect CLEVR accuracies that are measured by the usual methodology, as well as to contribute to the literature on systematic generalization. The specific aspect of systematicity that we analyze is the one exemplified in the previous paragraph: the ability to interpret known ways of referring to objects in arbitrary contexts. We focus on the \textit{matching referring expressions}
(see e.g. ``another cube that it is the same size as the brown cube'' in Figure \ref{fig:closure}) that require the object (or objects) to match another object in terms of a property, such as the size, the color, the material or the shape. 
We construct 7 CLOSURE tests with questions that highly overlap with the CLEVR ones and yet have zero probability under the CLEVR data distribution (see the Q3 in Figure \ref{fig:closure} for an example from one such test), aiming to cover the contexts in which property matching is not used to refer to objects in the original CLEVR. We call the resulting benchmark CLOSURE referring to the underlying idea of taking a closure (in the mathematical sense) of CLEVR questions under the operation of referring expression substitution and keeping those questions that are similar enough to the original ones. 

We evaluate a number of different CLEVR-trained models on CLOSURE, including end-to-end differentiable ones, like FiLM \citep{perez_learning_2017} and MAC \citep{hudson_compositional_2018}, and models using intermediate symbolic programs, like Neuro-Symbolic VQA (NS-VQA, \citet{yi_neural-symbolic_2018}) and the variety of Neural Module Networks (NMN, \citet{andreas_neural_2016}) proposed by \citet{johnson_inferring_2017}. We show that all aforementioned models often struggle on CLOSURE questions. For models using symbolic programs, such as NS-VQA and NMN, we observe a generalization gap in the performance of their neural sequence-to-sequence program generators \citep{sutskever_sequence_2014,bahdanau_neural_2015}. Furthermore, NMNs often exhibit poor generalization even when the ground-truth programs are provided. This result is remarkable given that the original motivation for NMNs is to decompose the model into components that can be recombined arbitrarily. To improve the NMN's generalization, we develop a new Vector-NMN module with a vector-valued (as opposed to tensor-valued) inputs and outputs. We show that the Vector-NMN modules perform much better than prior work when assembled in configurations that are different from the training ones. Lastly, we complement our zero-shot systematic generalization analysis with a few-shot transfer learning study and contrast the few-shot adaptation behavior of models with and without symbolic programs. 

\section{CLOSURE: A Systematic Generalization Benchmark for CLEVR}
\label{sec:closure}

\paragraph{Analysis of CLEVR} The key source of diversity and complexity in CLEVR questions is how objects of interest are referred to. We  call a noun phrase a \textit{referring expression} (RE) when it refers to an object or a set of objects, themselves called \textit{referents}. 
We distinguish three kinds of REs that occur in CLEVR: simple REs, complex REs and logical REs. A \textit{simple RE} is a noun that is (optionally) modified by one or more adjective, e.g. ``the big red cube'' or ``yellow shiny spheres''.
%\begin{align}
%    \text{\closureex{the big red cube}} \\
%    \text{\closureex{yellow shiny spheres}}.
%\end{align}
In \textit{Complex REs}, a relative clause (in square brackets in all examples below) is used to modify the noun (possibly in addition to adjectives):
\begin{align}
    \label{eq:loc-ref}\text{\closureex{the cube [that is left of $\langle RE \rangle$]}}, \\
    \label{eq:sim-ref}\text{\closureex{big cubes [that are the same color as $\langle RE \rangle$]}}.
\end{align}
Here,  $\langle RE \rangle$ is the \textit{embedded RE}, which can be either simple or also complex. Complex REs in CLEVR can be \textit{spatial} (Example \ref{eq:loc-ref}) or rely on matching objects' properties (Example \ref{eq:sim-ref}). We will call the latter \textit{matching REs}. The RE's type is determined by whether a \textit{spatial predicate} (``is left of $\langle RE \rangle$'', ``is right of $\langle RE \rangle$'', ...) or a \textit{matching predicate} (``is the same size as $\langle RE \rangle$'', ``is the same color as $\langle RE \rangle$'', ...) is used to construct the relative clause.\footnote{Note that the original paper by \citep{johnson_clevr:_2016} these are called ``spatial relationships'' and ``same-attribute relationships''}  In \textit{Logical REs}, two REs (Example \ref{eq:or-ref}) or two prepositional phrases (Example \ref{eq:and-ref}) are combined using ``and'' or ``or'':
\begin{align}
    \textrm{\label{eq:or-ref} } \text{\closureex{[balls] or [red blocks behind the cylinder]}},\\
    \label{eq:and-ref} \text{\closureex{ a cylinder that is [left of the brown ball]}} \notag \\
    \text{\closureex{and [in front of the tiny block]}}
\end{align}

The second most important axis of variation in CLEVR is what kind of question is asked about the referents. CLEVR includes existence, counting, attribute and object comparison questions, see examples below:
\begin{itemize}
\item (existence) Is there a small cyan object? 
\item (counting) How many purple things are behind the cylinder?
\item (attribute) What material is the big ball? 
\item (comparison) Do the red thing and the big thing have the same shape? 
\end{itemize}
In existence, counting and attribute questions there is one top-level RE, whereas comparison questions contain two top-level REs.  

\paragraph{CLOSURE questions} We have constructed the CLOSURE dataset by generating new CLEVR-like questions with matching predicates. To this end, we analyzed the composition of CLEVR and found a number of question templates in which a spatial predicate could be seamlessly substituted for a matching one. We focused on 7 cases where such substitution was possible and where it yielded questions that were not possible under CLEVR's original data distribution. Below, we describe and give examples of each of the resulting 7 CLOSURE tests. For a more technical explanation of the question generation procedure we refer the reader to Appendix \ref{app:closure}.

The {\tt\textbf{embed\_spa\_mat}} test contains existence questions with a matching RE that has an embedded spatial RE, e.g:
\begin{itemize}
   \item Is there a cylinder that is the same material as the object to the left of the blue thing?
\end{itemize}
Here, a spatial RE ``the object to the left of the blue thing'' is embedded in a matching RE ``a cylinder that is the same material ...''. Note, that in original CLEVR matching REs can only contain simple embedded REs. A closely related test is {\tt\textbf{embed\_mat\_spa}}, in which the top-level RE is spatial and the embedded one uses property matching:
\begin{itemize}
    \item \textrm{Is there a thing behind the cube that is the same color as the ball?}
\end{itemize}

In {\tt\textbf{compare\_mat}} and {\tt\textbf{compare\_mat\_spa}} tests we focus on models' ability to understand matching REs in comparison questions:
\begin{itemize}
    \item There is another small cylinder that is the same material as the small cyan cylinder; does it have the same color as the block?
    \item There is another cube that is the same material as the gray cube; does it have the same size as the metal thing to the right of the tiny gray cube?
\end{itemize}
The comparison questions in CLEVR only use spatial REs, hence {\tt compare\_mat} and {\tt compare\_mat\_spa} require models to recombine known constructs (that is the matching REs and the comparison questions) in a novel way. The two tests differ in whether the second RE is simple ({\tt compare\_mat}) or spatial ({\tt compare\_mat\_spa}).

The remaining three CLOSURE tests assess models' understanding of matching predicates in logical REs. The {\tt\textbf{or\_mat}} and {\tt\textbf{or\_mat\_spa}} questions require counting referents for a logical ``or'' of two REs, one of which uses property matching:
\begin{itemize}
    \item How many things are cubes or cylinders that are the same size as the red object?
    \item How many things are objects that are in front of the blue thing or small metallic things that are the same color as the rubber block?
\end{itemize}
The {\tt or\_mat\_spa} test differs from the {\tt or\_mat} one in that the second RE is also a complex one. The {\tt\textbf{and\_mat\_spa}} test contains attribute questions in which the RE involves a logical ``and'' of a spatial and a matching predicate:
\begin{itemize}
    \item What is the color of the thing that is to the left of the red cylinder and is the same size as the red block?
\end{itemize}
All the three tests presented above contain questions that are impossible under CLEVR's original data distribution, as logical REs in CLEVR only employ spatial predicates.

To construct CLOSURE questions and to compute the ground-truth answers we generated symbolic programs in the functional domain-specific language (DSL) that is CLEVR is based on. See Figure \ref{fig:programs} for example programs P1 and P2. Another usecase for groundtruth programs is to bootstrap learning in models that internally use programs as representations of questions' meanings, such e.g. NMNs. We refer the reader to Appendix \ref{app:dsl} for a description of the DSL and a detailed analysis of how CLOSURE programs different from the CLEVR ones.

\paragraph{Dataset statistics} We release  %\footnote{https://zenodo.org/record/3634090\#.Xjc3R3VKi90} 
a validation set, a test set and a small training set for few-shot learning investigations. 
The validation and test sets contain 3600 questions about the respective non-overlapping subsets of CLEVR validation images per each CLOSURE test.
The training questions (36 per CLOSURE test) are asked about training images from CLEVR. The data is available at https://zenodo.org/record/3634090\#.Xjc3R3VKi90

\section{Models}
\label{sec:models}
%A large number of models for the CLEVR task have been recently proposed, and it would be impossible for us to evaluate all of them. We therefore choose several models that vary in how CLEVR-specific their design is, aiming to cover the whole spectrum of ``CLEVR-awareness'' that such models possess. 
We experiments with a number of existing models, as well as with a novel Vector-NMN neural module that we employ in the context of the NMN paradigm.  Throughout this section we use capital letters for matrix- or tensor-shaped parameters of all models and small letters for the vector-valued ones. The symbols in equations that are not otherwise defined stand for trainable parameters (in particular this covers weight matrices $W, W_1, U, \ldots$ and bias vectors $b, b_1, \ldots$). We use $*$ to denote convolution as well as to inform the reader that the symbols on the left and right sides of the operator are a 4D and a 3D tensor respectively. $\odot$ and $ \oplus$ are used to denote feature-wise multiplication and addition for the case where one argument is a 3D tensor and another is a vector. The respective operation is applied independently to all sub-vectors of the tensor-valued argument obtained by fixing its first two indices (the approach known as ``broadcasting''). We use square brackets $\left[x;y\right]$ to denote tensor concatenation performed along the last axis.

\subsection{End-to-End Differentiable Models}

The most generic method that we consider is Feature-wise Linear Modulation (\textbf{FiLM}) by \citet{perez_film:_2017}. In this approach, an LSTM recurrent network transforms the question $q$ into biases $\beta$ and element-wise multipliers $\gamma$: $[\gamma; \beta ] = W \cdot LSTM(q) + b$. These so-called FiLM coefficients are then applied in the blocks of a deep residual convolutional network~\citep{he_deep_2016}. A FiLM-ed residual block takes a tensor-valued input $h_{in}$ and performs the following computation upon it:
\begin{align}
\tilde{h} = BN(W_2 * R(W_1 * h_{in} \oplus b_1) \oplus b_2), \\
h_{out} = R( h_{in} + \gamma \odot \tilde{h} \oplus \beta),
\end{align}
where $R$ stands for the Rectified Linear Unit, $BN$ denotes batch normalization \citep{ioffe_batch_2015}. Several such blocks are stacked together and applied to a 3D feature tensor $h_x$ that is produced by several layers of convolutions, some of them pretrained. The FiLM-ed network thus processes the input image $x$ in a manner that is modulated by the question $q$. Despite its simplicity, FiLM achieves a remarkably high reported accuracy of 97.7\% on the CLEVR task. A more advanced model that we include in our evaluation is Memory-Attention-Composition (\textbf{MAC}, see Appendix \ref{app:mac_model} for description) by \citet{hudson_compositional_2018}. 

%A more advanced model that we include in our evaluation is Memory-Attention-Composition (\textbf{MAC}) by \citet{hudson_compositional_2018}. In the MAC approach, the input and control components of the model first produce a sequence of control vectors $c_i$ from the question $q$. A visual attention component (called the read unit in the original paper) is then recurrently applied to a preprocessed version $h_x$ of the image $x$. The $i$-th application of the read unit is conditioned on the respective control vector $c_i$ and on a memory $m_i$ of the unit's outputs at the previous steps:
%\begin{align}
%    r_i = read\_unit(h_x, c_i, m_{i - 1}), \\
%    m_i = memory\_unit(r_i, m_{i-1}).
%\end{align}
%Such read operations and memory updates are performed for $T$ steps, after which the last memory vector $m_T$ and a question representation $q$ are concatenated and passed to the classifier. Different versions of the MAC model reach near-perfect 98.9-99.4\% performance on CLEVR.

\subsection{Modular and Symbolic Approaches}

In addition to the end-to-end differentiable models, we experiment with methods based on intermediate structured symbolic meaning representations. We adhere to the common practice of using programs expressed in the CLEVR DSL as such representations, although in principle logical formulae or other formalisms from the field of formal semantics could be used for this purpose. If a symbolic \textit{execution engine} for the programs is available, the task of VQA can be reduced to parsing the question and the image into a program and a symbolic scene representation respectively. This has already been proposed by \citet{yi_neural-symbolic_2018} under the name Neural-Symbolic VQA (\textbf{NS-VQA}) with a reported CLEVR accuracy of 99.8\%. This excellent performance, however, is achieved by relying heavily on the prior knowledge about the task, meaning that applying NS-VQA in conditions other than CLEVR could require significantly more adaptation and data collection than needed for the more generic methods, such as FiLM and MAC.

Intermediate symbolic programs can also be used without apriori knowledge of the semantics of the symbols, in which case the execution engine for the programs is either fully or partially learned. In the Neural Module Network (\textbf{NMN}) paradigm, proposed by \citet{andreas_neural_2016}, the meanings of symbols are represented in the form of trainable neural modules. Given a program, the modules that correspond to the program's symbols are retrieved and composed following the program's structure. Formally, a program in CLEVR DSL can be represented as a $(P, L, R)$ triple, where $P=(p_1, p_2, \ldots, p_T)$ is the sequence of function tokens\footnote{In this work we treat composite functions like e.g. ``filter\_color[brown]'' as standalone ones, not as ``filter\_color'' parameterized by ``brown''.}, $L=(l_1, \ldots, l_T)$ and $R=(r_1, \ldots, r_T)$ are the indices of the left and right arguments for each function call respectively (some DSL functions only take one or zero arguments, in which case the respective $r_i$ and $l_i$ are undefined). Using this formalism, a step of the NMN computation can be expressed as follows:
\begin{align}
h_i = 
\begin{cases}
M_{p_i}(h_x), & arity(p_i) = 0, \\
M_{p_i}(h_x, h_{l_i}), & arity(p_i) = 1, \\
M_{p_i}(h_x, h_{l_i}, h_{r_i}), & arity(p_i) = 2.
\end{cases} 
\end{align}
Here, $M_{p_i}$ is the neural module corresponding to the function token $p_i$ and $h_i$ is its output, while $h_x$ is a tensor representing the image.
Similar to MAC, the output $h_T$ of the last module is fed to the classifier, after which the modules are jointly trained by backpropagating the classifier's loss. To produce the programs for NMN training and usage a \textit{program generator} can be pretrained with a small seed set of (question, program)-pairs and then fine-tuned, e.g. with REINFORCE \citep{hu_learning_2017,johnson_inferring_2017}, on the rest of the dataset, using only (image, question, answer)-triplets as supervision. The programs produced by such a program generator can then be used at test time, meaning that after training the complete model takes the same inputs as end-to-end continuous models, such as FiLM and MAC. 

%A number of NMN-based approaches have been proposed for the CLEVR task, including those where different modules perform different operations (e.g. the module corresponding to logical ``and'' might compute an element-wise maximum of two vectors \citep{hu_learning_2017,mascharka_transparency_2018}), and those where all modules perform similar computations but use different parameters \citep{johnson_inferring_2017}. We focus on the latter variety of NMNs, since such models rely less on the domain knowledge and thus complement well the NS-VQA approach in our evaluation. In both cases, a \textit{program generator} can be pretrained with a small seed set of (question, program)-pairs and then fine-tuned, e.g. with REINFORCE \citep{hu_learning_2017,johnson_inferring_2017}, on the rest of the dataset, using only (image, question, answer)-triplets as supervision. The programs produced by such a program generator can then be used at test time, meaning that after training the complete model takes the same inputs as end-to-end continuous models, such as FiLM and MAC.

The first NMN model that we consider is the one proposed by \citet{johnson_inferring_2017}, in which residual blocks \citep{he_deep_2016} are used as neural modules $M_{p_i}$. For example, modules corresponding to functions of arity 2, (such as e.g. ``and'', ``equal\_color'', etc.), perform the following computation in their approach:
\begin{align}
    \tilde{h} = R(W_1 * \left[h_{l_i}; h_{r_i}\right]), \\
    h_i = R(W_3 * R(W_2 * \tilde{h} \oplus b_2) \oplus b_3 + \tilde{h}).
\end{align}
Note that the module described above does not use the image representation $h_x$ as an input; only the $M_{scene}$ module---the root node in all CLEVR programs---does so.

Our preliminary experiments showed that such modules often perform much worse when assembled in novel combinations. We hypothesized that this could be due to the fact that high-capacity 3D tensors $h_i$ are used in this model as the interface between modules. In order to test this hypothesis, we have designed a new module with a lower-dimensional vector output. We will henceforth refer to the module by \citet{johnson_inferring_2017} and our new module as \textbf{Tensor-NMN} and \textbf{Vector-NMN} respectively. The computation of our Vector-NMN is inspired by the FiLM approach to conditioning residual blocks on external inputs:
\begin{align}
\tilde{h}_1 = R(U_1 * (\gamma_1 \odot h_x \oplus \beta_1)),
\label{eq:vector1}\\
\tilde{h}_2 = R(U_2 * (\gamma_2 \odot \tilde{h}_1 \oplus \beta_2) + h_x), 
\label{eq:vector2}\\
h_{i} = \textrm{maxpool}(\tilde{h}_2),
\end{align}
where ``maxpool'' denotes max pooling of the 3D-tensor across all locations. Note that each Vector-NMN module also takes the image feature tensor $h_x$ as the input, unlike Tensor-NMN. The above equations describe a 1-block version of Vector-NMN, but in general several FiLM-ed residual blocks described by Equations \ref{eq:vector1} and \ref{eq:vector2} can be stacked prior to the pooling. The FiLM coefficients $\beta_1$, $\beta_2$, $\gamma_1$ and $\gamma_2$ are computed by 1-hidden-layer MLPs from the concatenation $h_{c} = \left[e(p_i); h_{l_i}; h_{r_i}\right]$ of the embedding $e(p_i)$ of the function token $p_i$ and the module inputs $h_{l_i}$ and $h_{r_i}$:
\begin{align}
\left[\beta_k, \tilde{\gamma_k} \right] = W^k_2 (R(W^k_1 h_{c} + b^k_1) + b^k_2),\\
\label{eq:tanh}
\gamma_k = 2 \tanh(\tilde{\gamma}_k) + 1.
\end{align} 
The $\tanh$ nonlinearity in Equation \ref{eq:tanh} was required to achieve stable training. Unlike in Tensor-NMN, the convolutional filters $U_1$ and $U_2$ are reused among all modules. To make this possible, we feed zero vectors instead of $h_{r_i}$ or $h_{l_i}$ when the function $p_i$ takes less than two inputs. 

\begin{figure}[t]
    \centering
    \includegraphics[width=\linewidth]{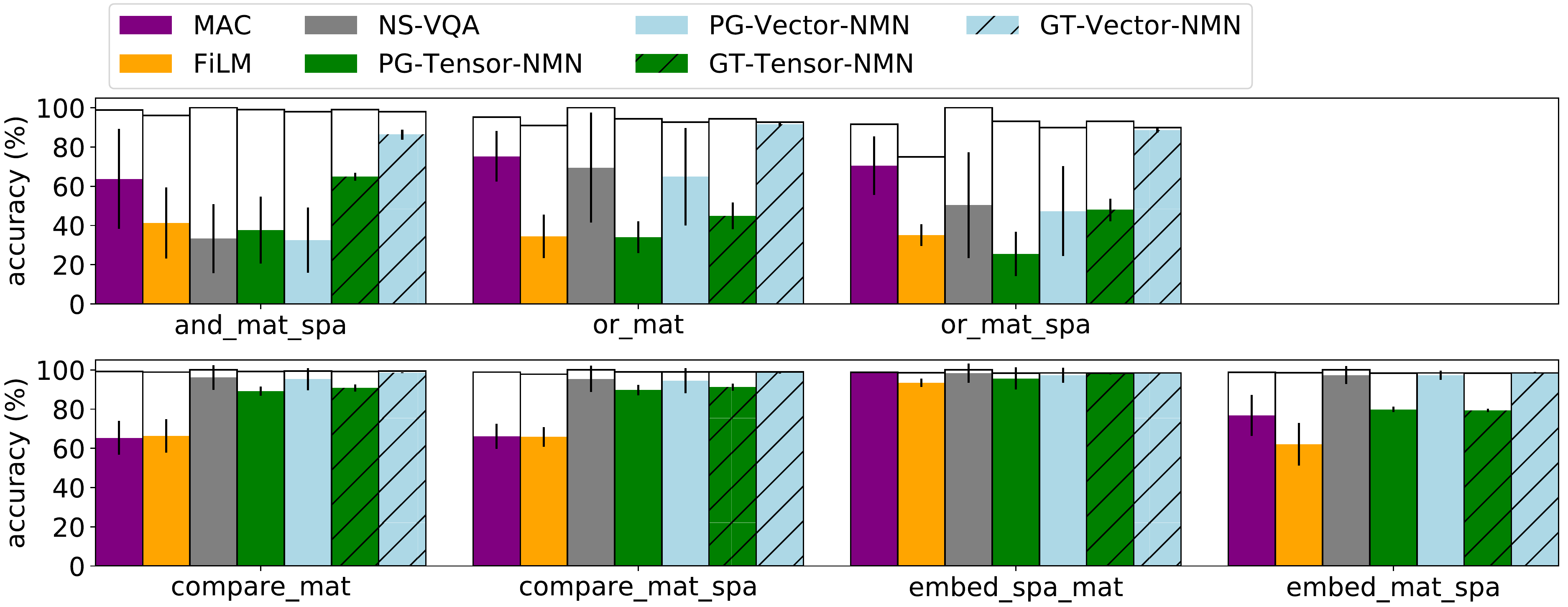}
    \caption{Zero-shot accuracy of all models on the 7 CLOSURE tests. For each model and test, the white bar in the background is the model's accuracy on the closest CLEVR questions. The hatching used for ``GT-...'' models indicates that we used the ground-truth programs at test time.}
    \label{fig:zeroshot}
\end{figure}
\begin{figure}[t]
    \centering
    \includegraphics[width=\linewidth]{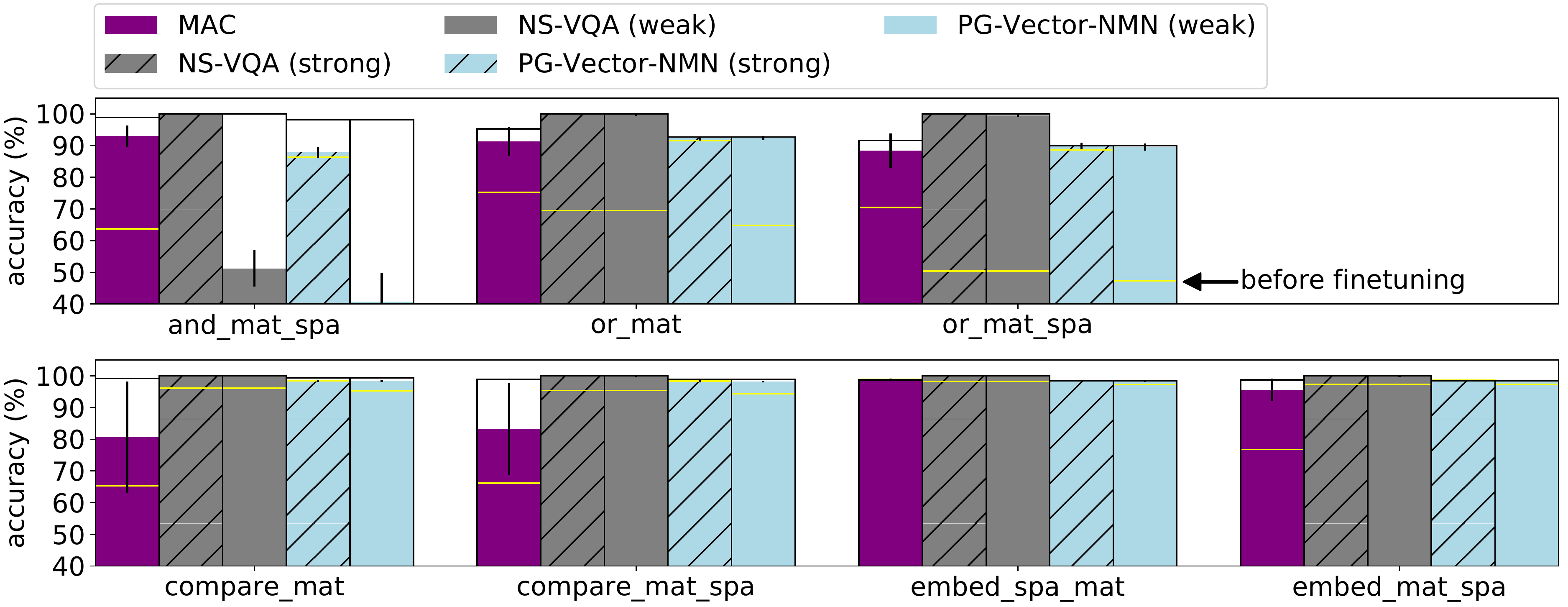}
    \caption{The accuracies for NS-VQA, PG-Vector-NMN and MAC after finetuning on 36 examples from each CLOSURE family. The background white bar is the model's accuracy on the closest CLEVR questions. The yellow horizontal line denotes the model's accuracy before fine-tuning. The hatching indicates the use of ground-truth programs at the fine-tuning stage.}
    \label{fig:fewshot}
\end{figure}

\section{Experiments}
We use the original implementation for FiLM and Tensor-NMN and train these models with the hyperparameter settings suggested by the authors. For the MAC model, we use a reimplementation 
that is close to the original one. We report results for a 2-block version of Vector-NMN, as our preliminary experiments on the original CLEVR dataset showed that it performs better than the 1-block version.
For all models that rely on symbolic programs, i.e. NS-VQA and the NMNs, we use a standard seq2seq model with an attention mechanism \citep{bahdanau_neural_2015} as the program generator. 
% Our preliminary investigations showed that this model generalizes better than the seq2seq models without attention~\citep{sutskever_sequence_2014,cho_learning_2014}, as used in \citep{johnson_inferring_2017}, and better than the seq2seq model used in the reference implementation of NS-VQA, in which the decoder does not take the attention outputs as inputs.
We report results for program generators trained with on the (question, program)-pairs from all 700k CLEVR examples.
% \footnote{Similarly to prior work we found that pretraining on 300-1000 ground-truth programs, followed by REINFORCE finetuning, is sufficient for near-perfect program generation performance.}.
In addition to evaluating the NMNs with the predicted programs, we also measure their performance when the ground-truth programs are given at test time. For the latter setting, we prepend \textbf{GT} to the model's name (as in GT-Vector-NMN), as opposed to prepending \textbf{PG} (as in PG-Vector-NMN). 
All numbers that we report are averages over 5 or 10 runs. Where relevant, we also report the standard deviation $\sigma$ in the form of $\pm \sigma$ or as a vertical black bar in figures. A detailed tabular presentation of the results is available in Appendix \ref{app:tabular}. For exact training settings and hyperparameters we refer the reader to Appendix \ref{app:training}.

\subsection{Zero-shot Generalization}
\label{sec:zero_shot}
In our first set of experiments, we assess zero-shot systematic generalization of models trained on CLEVR by measuring their performance on the CLOSURE tests. 
To put these results in context, for each model and test we measure the model's performance on validation questions from CLEVR that are most similar to the given test's questions.
For example, consider the {\tt embed\_spa\_mat} questions in which a spatial RE is embedded in a matching RE, such as ``Is there a big blue metal thing that is the same shape as the rubber object behind the blue shiny object?''. To establish a baseline for this test, we used existence questions where a spatial RE was embedded in another spatial RE, such as ``Is there a big blue metal thing that is behind the rubber object behind the blue shiny object?'' (notably, all models were accurate on $>98\%$ of such questions). The gap between the model's performance on baseline questions and the model's performance on a CLOSURE test is indicative of how systematic the model's generalization behavior is. 
% Indeed, the test- and model- dependent baseline scheme described above allows us to focus on the impact of combining known constructs in novel ways while controlling for other factors that influence the models' performance. For example, CLEVR-trained models typically perform worse on counting questions than on other question types, and hence we should expect lower accuracies on the {\tt or\_mat} and {\tt or\_mat\_spa} tests that require counting.  
% Dima right before EMNLP: add to appendix? but where?
See Appendix \ref{app:closure} for more details on baseline questions.

The results are reported in Figure \ref{fig:zeroshot}. One can clearly see that most models perform significantly worse on most CLOSURE tests compared to their accuracies on the respective baseline questions. A notable exception is {\tt embed\_spa\_mat}, on which all models perform quite well.  Among generic models, MAC consistently fares better than FiLM, albeit the former still loses 15\% to 35\% of its baseline accuracy in 6 out of 7 tests.
Surprisingly, the NS-VQA model, whose only learnable component is a program generator, generalizes outright badly on tests that involved logical references, and also shows significant deterioration compared to baseline questions on other tests.  This lack of systematic generalization in program generation also strongly affects performances of the two NMN models that we considered (PG-Vector-NMN and PG-Tensor-NMN). Interestingly, even given the ground-truth programs at test-time, Tensor-NMN modules perform worse on CLOSURE questions than on the respective baseline questions in 6 tests out of 7 (see GT-Tensor-NMN results). 
In contrast, our proposed Vector-NMN module generalizes much better, matching its baseline performance almost always, with a notable exception of the {\tt and\_mat\_spa} test. 
We conduct an additional investigation to better understand why Vector-NMN outperforms Vector-NMN. These extra results (Appendix \ref{app:vectornmn}) suggest that the max-pooling operation, which turns the Vector-NMN's output from a tensor into a vector, plays a key role in the module's improved generalization performance. 

\subsection{Few-shot transfer learning}
\label{sec:few_shot}
The results above show that inductive biases of existing models are often insufficient for zero-shot systematic generalization measured by CLOSURE. A natural question to ask in these circumstances is whether just a few examples would be sufficient to correct the models' extrapolation behavior. To answer this question, we finetune MAC, PG-Vector-NMN and NS-VQA models that are pretrained on CLEVR using 36 examples from each CLOSURE family, for a total of 252 new examples.
For PG-Vector-NMN and NS-VQA we consider two fine-tuning scenarios: one where the programs are provided for new examples and one where they are not given and must be inferred. To infer programs, we use a basic REINFORCE-based program induction approach \citep{johnson_inferring_2017,hu_learning_2017}.
We will refer to the two said scenarios as strong and weak supervision respectively. 
To get the best finetuning performance, we oversample 300 times the 252 training CLOSURE examples, add them to the CLEVR training set and train on the resulting mixed dataset. 
Just like in zero-shot experiments, we consider the model's performance on the closest CLEVR questions as the systematic generalization target.  

The few-shot results, reported in Figure \ref{fig:fewshot}, show that as few as 36 examples from each family can significantly improve CLOSURE performance for all models. A notable exception from this general observation is the {\tt and\_mat\_spa} question family, on which weakly supervised program induction for NS-VQA and PG-Vector-NMN most often did not work. We analyzed this case in detail and found that the appropriate programs for {\tt and\_mat\_spa} would typically have a very low probability, and hence were never sampled in our REINFORCE-based program search. 

The impact of 36 examples varies widely depending on the model and test. Models using symbolic programs reached the target performance in 6 tests out of 7.  On the contrary, for MAC a gap of $5\%$ to $20\%$ between its CLOSURE and target accuracies remained on all tests, except for the {\tt embed\_spa\_mat} test that MAC handled well even without fine-tuning. Notably, unlike the models relying on weakly-supervised program induction, MAC benefited greatly from fine-tuning on the challenging {\tt and\_mat\_spa} test. Besides, MAC's and PG-Vector-NMN's absolute performance on {\tt or\_mat} and {\tt or\_mat\_spa} are comparable.

\section{Related Work}
Several related generalization tests that were proposed for CLEVR and other VQA datasets differ from CLOSURE in what they aim to measure and/or how they were constructed. The Compositional Generalization Test (CoGenT) from the original paper by \citet{johnson_clevr:_2016} restricts the colors that cubes and cylinders can have in the training set images and inverts this restriction during the test time. By its design, CoGenT evaluates how robust a model is to a shift in the image distribution. On the contrary, in CLOSURE the image distribution remains the same at test time, but the question distribution changes to contain novel combinations of linguistic constructs from CLEVR. The generalization splits from the ShapeWorld platform \citep{kuhnle_shapeworld_2017} also focus on the difference in the distribution of images, not  questions. The CLEVR-Humans dataset was collected by having crowd workers ask questions about CLEVR images \citep{johnson_inferring_2017}. Some questions from this dataset require reasoning that is outside of the scope of CLEVR, such as e.g. quantification (``Are all the balls small?''). In contrast, CLOSURE requires models to recombine only the well-known reasoning primitives. A compositional C-VQA split was proposed for the VQA 1.0 dataset \citep{agrawal_c-vqa:_2017}. In C-VQA similar questions must have different answers when they appear in the training and test sets, yet the distributions of questions at training and testing remain similar, unlike CLOSURE. 

Perhaps the closest to our work is the SQOOP dataset and the study conducted on it by \citet{bahdanau_systematic_2019}. SQOOP features questions of the form ``Is there an X R of Y'', where X and Y are object words and R a spatial relation. The authors test whether models can answer all possible SQOOP questions after training on a subset that is defined by holding out most of the (X, Y) pairs. The methodology of that study is thus very similar to ours, however the specific nature of the generalization split is different. Similarly to our results, \citet{bahdanau_systematic_2019} report significant generalization gaps for a number of VQA models, with a notable exception of the Tensor-NMN, that generalized perfectly in their study when a tree-like layout was used to connect the modules. We believe our CLOSURE results are an important addition to the SQOOP ones.
The specific cause of the aforementioned discrepancies between the two studies is, however, an intriguing question for future work. 

As can be clearly seen from the performance of the NS-VQA model, much of the performance drop that we reported can be explained by insufficient systematicity of seq2seq models that we use for program generation. The SCAN dataset \citep{lake_generalization_2018} and the follow-up works \citep{loula_rearranging_2018,bastings_jump_2018} have recently brought much-needed attention to this important issue. Compared to SCAN, CLOSURE features richer and more natural-looking language, and hence can serve to validate the conclusions drawn in recent SCAN-based studies, e.g. \citep{russin_compositional_2019}. Concurrently to this work, \citep{hupkes_compositionality_2019} and \citep{keysers_measuring_2020} developed related benchmarks that also test systematicity of neural seq2seq transducers.

Prior work on NMNs features modules that output either attention maps \citep{andreas_neural_2016,hu_learning_2017,hu_explainable_2018} or feature tensors \citep{johnson_inferring_2017}. The model by \citep{mascharka_transparency_2018} combines modules with both attention- and tensor-valued outputs. Our Vector-NMN generalizes more systematically than its tensor-based predecessor by \citet{johnson_inferring_2017}, while inheriting that model's simplicity, generality and good CLEVR performance.

\section{Discussion} 
Our study shows that while models trained on CLEVR are very good at answering questions from CLEVR, their high performance quickly deteriorates when the question distribution features unfamiliar combinations of well-known primitives. We believe that this is an interesting finding, given that CLEVR puts VQA models in very favorable conditions: the training set is large and well-balanced and complex questions are well represented. 
%One could say that we took advantage of a gap in the CLEVR question distribution to make our point, yet we believe that natural language datasets collected under naturalistic conditions will only have more gaps like this.
% DIMA: the above commented out for EMNLP
%whereby certain expressions are typically used in a subset of contexts at the data collection time, but might occur in very different contexts once the data distribution shifts.

While in our zero-shot test of systematic generalization all models fare similarly badly, our few-shot learning study highlights important differences in their behavior. Given few examples, the program-based models either almost perfectly adapt to the target task or completely fail, depending on whether the right programs are found.  For end-to-end continuous models back-propagation is always effective in adapting them to the target examples, but the systematic generalization gap is often not fully bridged with a number of examples that is sufficient for the program-based models. An important context for this comparison is that program-based models require seed programs to jump-start the training and are later on constrained to the seed lexicon.  It would be highly desirable to combine the strengths of these two types of systems in one model without inheriting any of their limitations, a direction that we would like to explore in our future work.

% We hope that the CLOSURE benchmark will facilitate future work in a number of directions. First, our results suggest that parsing (program generation in our case) can limit systematic generalization of grounded language learning. CLOSURE can thus be used for testing  systematic generalization of neural parsers, complementing the SCAN benchmark and its variants. Besides parsing, further work on interchangeability of neural modules can be done using CLOSURE. While Vector-NMN improves upon prior work, it still generalizes suboptimally on the {\tt and\_mat\_spa} test.  Lastly, our test set construction methods can be adapted to natural data, yielding more insights and helping researchers make measurable progress towards learning-based models for grounded language understanding that generalize systematically. 

\bibliography{references}
\bibliographystyle{icml2020}

\newpage
\onecolumn
\appendix

\section{The CLEVR DSL and CLOSURE Programs Expressed In It}
\label{app:dsl}

The DSL functions that implement meanings of referring expressions operate on sets of objects, where each object is represented by the values of its four properties (shape, color, size and material) and its spatial coordinates. \emph{Filter} functions filter the input set of objects by the value of a property (e.g. ``filter\_color[brown]'', ``filter\_shape[cube]''), and \emph{relations} return a set of all other objects that are related to the given object. The two kinds of relations in the DSL correspond to the spatial and matching predicates that we discussed above. Namely, there are spatial relations: (``relate[left]'', ``relate[right]'', ...) and matching relations (``same\_shape'', ``same\_color'', ...). The subprograms that correspond to the REs consist of chained filters and relations, as well as set union and set intersection functions in the case of logical REs\footnote{The ``unique'' function also frequently appears in the subprograms that correspond to REs. Its only role is to raise an exception if its input is not a singleton set.}. The wider use of matching predicates that distinguishes CLOSURE questions from CLEVR ones translates into matching relations appearing in more diverse kinds of programs than in CLEVR. For example, in program P3 in Figure \ref{fig:programs} the relation ``same\_size'' appears in the same program with the function ``equal\_color'', a combination that would not be possible in CLEVR. Hence, to succeed on CLOSURE, the NMN models have to learn modules which can be arbitrarily recombined with each other.

\section{Memory-Attention-Composition (MAC) Model} 
\label{app:mac_model} 

In the MAC approach by \citep{hudson_compositional_2018}, the input and control components of the model first produce a sequence of control vectors $c_i$ from the question $q$. A visual attention component (called the read unit in the original paper) is then recurrently applied to a preprocessed version $h_x$ of the image $x$. The $i$-th application of the read unit is conditioned on the respective control vector $c_i$ and on a memory $m_i$ of the unit's outputs at the previous steps:
%\begin{align}
%    r_i = read\_unit(h_x, c_i, m_{i - 1}), \\
%    m_i = memory\_unit(r_i, m_{i-1}).
%\end{align}
%Such read operations and memory updates are performed for $T$ steps, after which the last memory vector $m_T$ and a question representation $q$ are concatenated and passed to the classifier. Different versions of the MAC model reach near-perfect 98.9-99.4\% performance on CLEVR.

\section{Tabular CLOSURE Results}
\label{app:tabular}
See Table \ref{tbl:zeroshot} for zero-shot results, Table \ref{tbl:fewshot} for few-shot results, and Table \ref{tbl:baseline} for the target performances for all CLOSURE tests.

%%%%%%%%%%%%%%%%%%%%%%%%%%%%%%%%%%%%%%% ZEROSHOT %%%%%%%%%%%%%%%%%%%%%%%%%%%%%%%%%%%%%%%%%%%

\begin{table}[h]
\begin{subfigure}[b]{\textwidth}
\centering
\begin{tabular}{llll}
\toprule
model                          & {\tt and\_mat\_spa}   & {\tt or\_mat}         & {\tt or\_mat\_spa}   \\
\midrule
FiLM                           & $41.4\pm18$     & $34.4\pm11$     & $35.2\pm5.5$   \\
MAC                            & $63.7\pm25$     & $75.2\pm13$     & $70.4\pm15$    \\
NS-VQA                         & $33.3\pm18$     & $69.5\pm28$     & $50.4\pm27$    \\
PG-Tensor-NMN                  & $37.6\pm17$     & $34.1\pm8.1$    & $25.5\pm11$    \\
PG-Vector-NMN                  & $32.5\pm17$     & $64.9\pm25$     & $47.4\pm23$    \\
GT-Tensor-NMN                  & $64.9\pm2$      & $44.8\pm6.8$    & $47.9\pm5.8$   \\
GT-Vector-NMN                  & $86.3\pm2.5$    & $91.5\pm0.77$   & $88.6\pm1.2$   \\
\end{tabular}
\end{subfigure}
\vspace{0.5cm}
\begin{subfigure}[b]{\textwidth}
\centering
\begin{tabular}{lllll}
\toprule
model                          & {\tt embed\_spa\_mat} & \tt{embed\_mat\_spa} & {\tt compare\_mat}    & {\tt compare\_mat\_spa}\\
\midrule
FiLM                           & $93.4\pm2$      & $62\pm11$       & $66.2\pm8.5$    & $65.8\pm5$     \\
MAC                            & $99\pm0.15$     & $76.8\pm11$     & $65.3\pm8.6$    & $66.2\pm6.4$   \\
NS-VQA                         & $98.3\pm5$      & $97.3\pm4.6$    & $96.1\pm6.2$    & $95.4\pm6.7$   \\
PG-Tensor-NMN                  & $95.6\pm5.6$    & $79.8\pm1.4$    & $89.2\pm2.3$    & $89.8\pm2.7$   \\
PG-Vector-NMN                  & $97.2\pm3.9$    & $97.3\pm2.4$    & $95.3\pm5.7$    & $94.4\pm6.4$   \\
GT-Tensor-NMN                  & $98.1\pm0.38$   & $79.3\pm0.83$   & $90.7\pm1.8$    & $91.2\pm1.9$   \\
GT-Vector-NMN                  & $98.5\pm0.13$   & $98.7\pm0.19$   & $98.5\pm0.17$   & $98.4\pm0.3$   \\
\bottomrule
\end{tabular}
\end{subfigure}
\caption{Zero-shot performance of all models on CLOSURE tests. Reported is the mean accuracy and its standard deviation.}
\label{tbl:zeroshot}
\end{table}

%%%%%%%%%%%%%%%%%%%%%%%%%%%%%%%%%%%%%%% FEWSHOT %%%%%%%%%%%%%%%%%%%%%%%%%%%%%%%%%%%%%%%%%%%

\begin{table}[h]
\begin{subfigure}[b]{\textwidth}
\centering
\begin{tabular}{llll}
\toprule
model                          & {\tt and\_mat\_spa}   & {\tt or\_mat}         & {\tt or\_mat\_spa}   \\
\midrule
NS-VQA (strong)                & 100             & $100\pm0.026$   & $100\pm0.079$  \\
NS-VQA (weak)                  & $51.2\pm5.8$    & $99.8\pm0.38$   & $99.5\pm0.42$  \\
PG-Vector-NMN (strong)         & $87.8\pm1.7$    & $92.3\pm0.68$   & $89.9\pm1$     \\
PG-Vector-NMN(weak)            & $40.8\pm8.9$    & $92.4\pm0.65$   & $89.5\pm1.1$   \\
MAC                            & $93\pm3.3$      & $91.3\pm4.6$    & $88.4\pm5.5$   \\
\end{tabular}
%\caption{AND, OR results}
\end{subfigure}
\vspace{0.5cm}
\begin{subfigure}[b]{\textwidth}
\centering
\begin{tabular}{lllll}
\toprule
model                          & {\tt embed\_spa\_mat} & \tt{embed\_mat\_spa} & {\tt compare\_mat}    & {\tt compare\_mat\_spa}\\
\midrule
NS-VQA (strong)                & 100             & 100             & 100             & 100            \\
NS-VQA (weak)                  & $100\pm0.026$   & $99.9\pm0.24$   & $100\pm0.034$   & $99.8\pm0.24$  \\
PG-Vector-NMN (strong)         & $98.5\pm0.21$   & $98.5\pm0.26$   & $98.3\pm0.29$   & $98.3\pm0.31$  \\
PG-Vector-NMN(weak)            & $98.4\pm0.26$   & $98.5\pm0.17$   & $98.4\pm0.37$   & $98.2\pm0.29$  \\
MAC                            & $99\pm0.12$     & $95.6\pm3.6$    & $80.7\pm18$     & $83.3\pm15$    \\
\bottomrule
\end{tabular}
\end{subfigure}
\caption{Performance of several models on CLOSURE tests after finetuning on 252 CLOSURE examples, 36 from each family. Reported is the mean accuracy and its standard deviation.}
\label{tbl:fewshot}
\end{table}

%%%%%%%%%%%%%%%%%%%%%%%%%%%%%%%%%%%%%%% BASELINES %%%%%%%%%%%%%%%%%%%%%%%%%%%%%%%%%%%%%%%%%%%

\begin{table}[h]
\begin{subfigure}[b]{\textwidth}
\centering
\begin{tabular}{llll}
\toprule
model                          & {\tt and\_mat\_spa}   & {\tt or\_mat}         & {\tt or\_mat\_spa}   \\
\midrule
MAC                            & $98.9\pm0.35$   & $95.3\pm0.62$   & $91.6\pm2.6$   \\
FiLM                           & $96.1\pm1.6$    & $91\pm0.99$     & $75\pm2.1$     \\
NS-VQA                         & 100             & $100\pm0.0088$  & 100            \\
PG-Tensor-NMN                  & $99\pm0.3$      & $94.4\pm0.67$   & $93.1\pm0.62$  \\
PG-Vector-NMN                  & $98.1\pm0.33$   & $92.7\pm1.1$    & $89.9\pm1.3$   \\
GT-Tensor-NMN                  & $99\pm0.3$      & $94.4\pm0.67$   & $93.1\pm0.62$  \\
GT-Vector-NMN                  & $98.1\pm0.33$   & $92.7\pm1.1$    & $89.9\pm1.3$   \\
\end{tabular}
\end{subfigure}
\begin{subfigure}[b]{\textwidth}
\centering
\begin{tabular}{lllll}
\toprule
model                          & {\tt embed\_spa\_mat} & \tt{embed\_mat\_spa} & {\tt compare\_mat}    & {\tt compare\_mat\_spa}\\
\midrule
MAC                            & $98.8\pm0.56$   & $98.8\pm0.56$   & $99.2\pm0.15$   & $98.9\pm0.27$  \\
FiLM                           & $98.5\pm0.54$   & $98.5\pm0.54$   & $98.9\pm0.39$   & $97.7\pm0.76$  \\
NS-VQA                         & 100             & 100             & 100             & 100            \\
PG-Tensor-NMN                  & $98.3\pm0.81$   & $98.3\pm0.81$   & $99.2\pm0.32$   & $98.9\pm0.17$  \\
PG-Vector-NMN                  & $98.5\pm0.53$   & $98.5\pm0.53$   & $99.4\pm0.17$   & $98.9\pm0.24$  \\
GT-Tensor-NMN                  & $98.3\pm0.81$   & $98.3\pm0.81$   & $99.2\pm0.32$   & $98.9\pm0.17$  \\
GT-Vector-NMN                  & $98.5\pm0.53$   & $98.5\pm0.53$   & $99.4\pm0.17$   & $98.9\pm0.24$  \\
\bottomrule
\end{tabular}
\end{subfigure}
\caption{Models' performance on CLEVR questions that are the closest to a given CLOSURE test (see Appendix \ref{app:closure} for additional details regarding the closest questions for {\tt or\_mat} and {\tt or\_mat\_spa}). Reported is the mean accuracy and its standard deviation.}
\label{tbl:baseline}
\end{table}

\section{Results for the original MAC implementation}

To validate our conclusions, we also tried to measure performance of the reference MAC implementation by \citet{hudson_compositional_2018} on CLOSURE tests. We ran the codes 10 times using the default configuration. The results are reported in Table \ref{tbl:mac_orig}. While some accuracy differences are statistical significant (e.g. {\tt or\_mat}), these results would lead to same conclusions as the ones that we obtained by using the approximate reimplementation by \citet{bahdanau_systematic_2019}. The latter was more convenient for us to use because it is written using the framework that we use for all other models, namely PyTorch.

%%%%%%%%%%%%%%%%%%%%%%%%%%%%%%%%%%%%%%% ORIGINAL MAC %%%%%%%%%%%%%%%%%%%%%%%%%%%%%%%%%%%%%%%%%%%

\begin{table}
\begin{subfigure}[b]{\textwidth}
\centering
\begin{tabular}{llll}
\toprule
model                          & {\tt and\_mat\_spa}   & {\tt or\_mat}         & {\tt or\_mat\_spa}   \\
\midrule
reimpl. by \citep{bahdanau_systematic_2019}            & $63.7\pm25$     & $75.2\pm13$     & $70.4\pm15$    \\
original by \citep{hudson_compositional_2018} & $77.1\pm11$     & $57.8\pm7.7$    & $55.3\pm6.3$   \\
\end{tabular}
\end{subfigure}
\begin{subfigure}[b]{\textwidth}
\centering
\begin{tabular}{lllll}
\toprule
model                          & {\tt embed\_spa\_mat} & \tt{embed\_mat\_spa} & {\tt compare\_mat}    & {\tt compare\_mat\_spa}\\
\midrule
reimpl. by \citep{bahdanau_systematic_2019}            & $99\pm0.15$     & $76.8\pm11$     & $65.3\pm8.6$    & $66.2\pm6.4$   \\
original by \citep{hudson_compositional_2018} & $98.6\pm0.89$   & $69.5\pm7.5$    & $65.1\pm5.3$    & $64.3\pm3.1$   \\
\bottomrule
\end{tabular}
\end{subfigure}
\caption{Comparison between CLOSURE accuracies of two MAC implementation, the original one by \citet{hudson_compositional_2018}, and the reimplementation by \citet{bahdanau_systematic_2019}}
\label{tbl:mac_orig}
\end{table}

\section{CLEVR results}
In Table \ref{tbl:clevr} we report the accuracies on the CLEVR validation test for all the models that we consider in this study (including the ones introduced below in Appendix \ref{app:vectornmn}).
\begin{table}
\centering
\begin{tabular}{cc}
\toprule
model                          & CLEVR accuracy \\
\midrule
MAC                            & $98.5\pm0.11$  \\
FiLM                           & $97\pm0.56$    \\
NS-VQA                         & $100\pm0.00021$\\
PG-Tensor-NMN                  & $97.9\pm0.052$ \\
PG-Vector-NMN                  & $98\pm0.066$   \\
GT-Tensor-NMN                  & $97.9\pm0.052$ \\
GT-Vector-NMN                  & $98\pm0.066$   \\
GT-Tensor-NMN-Shortcuts        & $98\pm0.079$   \\
GT-Tensor-NMN-FiLM             & $94.8\pm1.7$   \\
\bottomrule
\end{tabular}
\caption{CLEVR validation accuracies for all models}
\label{tbl:clevr}
\end{table}

\section{Further details on CLOSURE and baseline questions}
\label{app:closure}
%Indeed, the test- and model- dependent baseline scheme described above allows us to focus on the impact of combining known constructs in novel ways while controlling for other factors that influence the models' performance. For example, CLEVR-trained models typically perform worse on counting questions than on other question types, and hence we should expect lower accuracies on the {\tt or\_mat} and {\tt or\_mat\_spa} tests that require counting.
The exact CLOSURE templates can be found in Figure \ref{tbl:templates}. The templates for the baseline questions, that is CLEVR questions that are most similar to CLOSURE ones, can be found in Figure \ref{tbl:baseline questions}. The slots in the templates should be understood as follows:
\begin{itemize}
    \item the slots $\langle$A$\rangle$ and $\langle$Q$\rangle$ will be filled by names of CLEVR properties, such as ``shape'', ``size'', ``color'' and ``material''
    \item the slots $\langle$Z$\rangle$, $\langle$C$\rangle$, $\langle$M$\rangle$ and $\langle$S$\rangle$ will be either left blank or filled by words that characterize objects' properties, such as ``big'', ``small'', ``yellow'', ``rubber'', ``cube'', etc.
    \item the $\langle$R$\rangle$ slots will be filled with spatial words, such as ``left of'', ``right of'', ``behind'' or ``in front of''.
\end{itemize}
In some templates fragments of text are in square brackets; those are optional and will be discarded with 50\% probability.

We generate the questions from the templates in two stages. At the first stage, we fill the $\langle$A$\rangle$ and $\langle$Q$\rangle$ slots, and at the second stage we use CLEVR question generation code to fill the rest. The two-stage procedure is required because the original generation engine does not support $\langle$A$\rangle$ and $\langle$Q$\rangle$ slots; instead, CLEVR authors wrote unique templates for each property that is queried (the purpose of $\langle$Q$\rangle$) or used to refer to objects (the purpose of $\langle$A$\rangle$). 

We used CLEVR generation code mostly as is, with the exception of two important modifications. First, we restricted the answers of counting questions to be either 1, 2 or 3, and also applied more aggressive rejection sampling to make these answers equally likely. We did so because the distribution of answers to counting questions in CLEVR is skewed, and answers of 4 and more are very unlikely. Instead of trying to replicate the original skewed answer distribution, we chose to enforce uniformity among those answers (that is 1, 2 and 3) that do have a significant probability in CLEVR. We did not allow 0 as the answer because due to implementation details, CLEVR questions that contain logical ``or'' and a complex spatial RE never have 0 as the answer. Overall, with our modifications to generation of counting questions we tried to put side the irrelevant confounding factors and focus on the impact of replacing a spatial RE with a matching one. To compute the appropriate target performance for {\tt or\_mat} and {\tt or\_mat\_spa}, we generated new questions from the closest original CLEVR templates but using the modified version of the generation code.

The second modification concerns the question degeneracy check that is described in the appendix of \citep{johnson_clevr:_2016}. In the reference question generation code it is only applied to programs with spatial relations. We modified the code to also apply the degeneracy check to matching relations. 

A minor issue with {\tt compare\_mat} and {\tt compare\_mat\_spa} questions is that the word ``another'' can be used in cases where it is not required, e.g. ``... another cube that is the same size as the sphere''. CLEVR generation code removes ``another'' in such cases, but we found it hard to extend this feature to CLOSURE questions in a maintainable way. In our preliminary experiments we found the proper handling of ``another'' does not change results of zero-shot experiments. 
We also experimented with removing the word ``is'' from ``... and is the same $\langle$A$\rangle$ ...'' in the {\tt and\_mat\_spa} template to make it more similar to the closest CLEVR questions, in which ``and'' combines prepositional phrases (i.e. ``... that is left of the cube and right of the sphere''). Likewise, we saw no influence on the zero-shot results. 

\begin{figure}
\begin{itemize}
\item {\tt\textbf{embed\_spa\_mat}}
Is there a $\langle$Z$\rangle$ $\langle$C$\rangle$ $\langle$M$\rangle$ $\langle$S$\rangle$ that is the same $\langle$A$\rangle$ as the $\langle$Z2$\rangle$ $\langle$C2$\rangle$ $\langle$M2$\rangle$ $\langle$S2$\rangle$ $\langle$R$\rangle$ the $\langle$Z3$\rangle$ $\langle$C3$\rangle$ $\langle$M3$\rangle$ $\langle$S3$\rangle$?
\item {\tt\textbf{embed\_mat\_spa}}
Is there a $\langle$Z$\rangle$ $\langle$C$\rangle$ $\langle$M$\rangle$ $\langle$S$\rangle$ $\langle$R$\rangle$ the $\langle$Z2$\rangle$ $\langle$C2$\rangle$ $\langle$M2$\rangle$ $\langle$S2$\rangle$ that is the same $\langle$A$\rangle$ as $\langle$Z3$\rangle$ $\langle$C3$\rangle$ $\langle$M3$\rangle$ $\langle$S3$\rangle$?
\item {\tt\textbf{compare\_mat}}
There is another $\langle$Z$\rangle$ $\langle$C$\rangle$ $\langle$M$\rangle$ $\langle$S$\rangle$ that is the same $\langle$A$\rangle$ as the $\langle$Z2$\rangle$ $\langle$C2$\rangle$ $\langle$M2$\rangle$ $\langle$S2$\rangle$; does it have the same $\langle$Q$\rangle$ as the $\langle$Z3$\rangle$ $\langle$C3$\rangle$ $\langle$M3$\rangle$ $\langle$S3$\rangle$?
\item {\tt\textbf{compare\_mat\_spa}}
There is another $\langle$Z$\rangle$ $\langle$C$\rangle$ $\langle$M$\rangle$ $\langle$S$\rangle$ that is the same $\langle$A$\rangle$ as the $\langle$Z2$\rangle$ $\langle$C2$\rangle$ $\langle$M2$\rangle$ $\langle$S2$\rangle$; does it have the same $\langle$Q$\rangle$ as the $\langle$Z3$\rangle$ $\langle$C3$\rangle$ $\langle$M3$\rangle$ $\langle$S3$\rangle$ [that is] $\langle$R2$\rangle$ the $\langle$Z4$\rangle$ $\langle$C4$\rangle$ $\langle$M4$\rangle$ $\langle$S4$\rangle$?
\item {\tt\textbf{and\_mat\_spa}}
What is the $\langle$Q$\rangle$ of the $\langle$Z$\rangle$ $\langle$C$\rangle$ $\langle$M$\rangle$ $\langle$S$\rangle$ that is $\langle$R2$\rangle$ the $\langle$Z2$\rangle$ $\langle$C2$\rangle$ $\langle$M2$\rangle$ $\langle$S2$\rangle$ and is the same $\langle$A$\rangle$ as the $\langle$Z3$\rangle$ $\langle$C3$\rangle$ $\langle$M3$\rangle$ $\langle$S3$\rangle$?
\item {\tt\textbf{or\_mat}}
How many things are [either] $\langle$Z$\rangle$ $\langle$C$\rangle$ $\langle$M$\rangle$ $\langle$S$\rangle$s or $\langle$Z2$\rangle$ $\langle$C2$\rangle$ $\langle$M2$\rangle$ $\langle$S2$\rangle$s that are the same $\langle$A$\rangle$ as the $\langle$Z3$\rangle$ $\langle$C3$\rangle$ $\langle$M3$\rangle$ $\langle$S3$\rangle$?
\item {\tt\textbf{or\_mat\_spa}}
How many things are [either] $\langle$Z$\rangle$ $\langle$C$\rangle$ $\langle$M$\rangle$ $\langle$S$\rangle$s [that are] $\langle$R$\rangle$ the $\langle$Z2$\rangle$ $\langle$C2$\rangle$ $\langle$M2$\rangle$ $\langle$S2$\rangle$ or $\langle$Z3$\rangle$ $\langle$C3$\rangle$ $\langle$M3$\rangle$ $\langle$S3$\rangle$s that are the same $\langle$A$\rangle$ as the $\langle$Z4$\rangle$ $\langle$C4$\rangle$ $\langle$M4$\rangle$ $\langle$S4$\rangle$?
\end{itemize}
\caption{CLOSURE templates.}
\label{tbl:templates}
\end{figure}

\begin{figure}
\begin{itemize}
\item {\tt\textbf{embed\_spa\_mat}}
Is there a $\langle$Z$\rangle$ $\langle$C$\rangle$ $\langle$M$\rangle$ $\langle$S$\rangle$ [that is] $\langle$R$\rangle$ the $\langle$Z2$\rangle$ $\langle$C2$\rangle$ $\langle$M2$\rangle$ $\langle$S2$\rangle$ [that is] $\langle$R2$\rangle$ the $\langle$Z3$\rangle$ $\langle$C3$\rangle$ $\langle$M3$\rangle$ $\langle$S3$\rangle$?
\item {\tt\textbf{embed\_mat\_spa}}
Is there a $\langle$Z$\rangle$ $\langle$C$\rangle$ $\langle$M$\rangle$ $\langle$S$\rangle$ [that is] $\langle$R$\rangle$ the $\langle$Z2$\rangle$ $\langle$C2$\rangle$ $\langle$M2$\rangle$ $\langle$S2$\rangle$ [that is] $\langle$R2$\rangle$ the $\langle$Z3$\rangle$ $\langle$C3$\rangle$ $\langle$M3$\rangle$ $\langle$S3$\rangle$?
\item {\tt\textbf{compare\_mat}}
There is a $\langle$Z$\rangle$ $\langle$C$\rangle$ $\langle$M$\rangle$ $\langle$S$\rangle$ [that is] $\langle$R$\rangle$ the $\langle$Z2$\rangle$ $\langle$C2$\rangle$ $\langle$M2$\rangle$ $\langle$S2$\rangle$; does it have the same $\langle$Q$\rangle$ as the $\langle$Z3$\rangle$ $\langle$C3$\rangle$ $\langle$M3$\rangle$ $\langle$S3$\rangle$?
\item {\tt\textbf{compare\_mat\_spa}}
There is a $\langle$Z$\rangle$ $\langle$C$\rangle$ $\langle$M$\rangle$ $\langle$S$\rangle$ [that is] $\langle$R$\rangle$ the $\langle$Z2$\rangle$ $\langle$C2$\rangle$ $\langle$M2$\rangle$ $\langle$S2$\rangle$; does it have the same $\langle$Q$\rangle$ as the $\langle$Z3$\rangle$ $\langle$C3$\rangle$ $\langle$M3$\rangle$ $\langle$S3$\rangle$ [that is] $\langle$R2$\rangle$ the $\langle$Z4$\rangle$ $\langle$C4$\rangle$ $\langle$M4$\rangle$ $\langle$S4$\rangle$?
\item {\tt\textbf{and\_mat\_spa}}
What is the $\langle$Q$\rangle$ of the $\langle$Z$\rangle$ $\langle$C$\rangle$ $\langle$M$\rangle$ $\langle$S$\rangle$ that is [both] $\langle$R$\rangle$ the $\langle$Z2$\rangle$ $\langle$C2$\rangle$ $\langle$M2$\rangle$ $\langle$S2$\rangle$ and $\langle$R2$\rangle$ the $\langle$Z3$\rangle$ $\langle$C3$\rangle$ $\langle$M3$\rangle$ $\langle$S3$\rangle$?
\item {\tt\textbf{or\_mat}}
How many things are [either] $\langle$Z$\rangle$ $\langle$C$\rangle$ $\langle$M$\rangle$ $\langle$S$\rangle$s or $\langle$Z2$\rangle$ $\langle$C2$\rangle$ $\langle$M2$\rangle$ $\langle$S2$\rangle$s [that are] $\langle$R$\rangle$ the $\langle$Z3$\rangle$ $\langle$C3$\rangle$ $\langle$M3$\rangle$ $\langle$S3$\rangle$?
\item {\tt\textbf{or\_mat\_spa}}
How many things are [either] $\langle$Z$\rangle$ $\langle$C$\rangle$ $\langle$M$\rangle$ $\langle$S$\rangle$s [that are] $\langle$R$\rangle$ the $\langle$Z2$\rangle$ $\langle$C2$\rangle$ $\langle$M2$\rangle$ $\langle$S2$\rangle$ or $\langle$Z3$\rangle$ $\langle$C3$\rangle$ $\langle$M3$\rangle$ $\langle$S3$\rangle$s [that are] $\langle$R2$\rangle$ the $\langle$Z4$\rangle$ $\langle$C4$\rangle$ $\langle$M4$\rangle$ $\langle$S4$\rangle$?
\end{itemize}
\caption{Baseline CLEVR templates.}
\label{tbl:baseline questions}
\end{figure}

\section{Training Details}
\label{app:training}
We train all models using the Adam optimizer \citep{kingma_adam:_2015} with the hyperparameters $\beta_1=0.9$, $\beta_2=0.999$, $\epsilon=10^{-8}$. The learning rate, the weight decay and the batch size varied across models, the exact values can be found in Table \ref{tbl:hyperparams}.

\begin{table}
    \centering
    \begin{tabular}{cccccc}
        \toprule 
        model & learning rate & weight decay & batch size & training iterations & \# of runs\\
        \midrule
        MAC & $10^{-4}$ & $10^{-5}$ & 64 & 1000000 & 10 \\
        FiLM & $3 \cdot 10^{-4}$ & $10^{-5}$ & 64 & 500000 & 5 \\
        Tensor-NMN & $10^{-4}$ & 0 & 64 & 500000 & 5\\
        Vector-NMN & $10^{-4}$ & 0 & 128 & 500000 & 10\\
        PG & $7 \cdot 10^{-4}$ & 0 & 64 & 100000 & 10\\
        PG (REINFORCE) & $10^{-5}$ & 0 & 128 & 200000 & 10\\
        \bottomrule
    \end{tabular}
    \caption{Key hyperparameters and training settings. ``PG`` stands for the program generator that we used in PG-Vector-NMN and PG-Tensor-NMN, as well as in the NS-VQA model. The ``PG (REINFORCE)'' row contains the hyperparameters that we used to fine-tune the program generator on CLOSURE examples. For explanation of other acronyms, see Section \ref{sec:models} in the main text.}
    \label{tbl:hyperparams}
\end{table}

\section{Vector-NMN}
\label{app:vectornmn}
\begin{figure}
\centering
\includegraphics[scale=0.5]{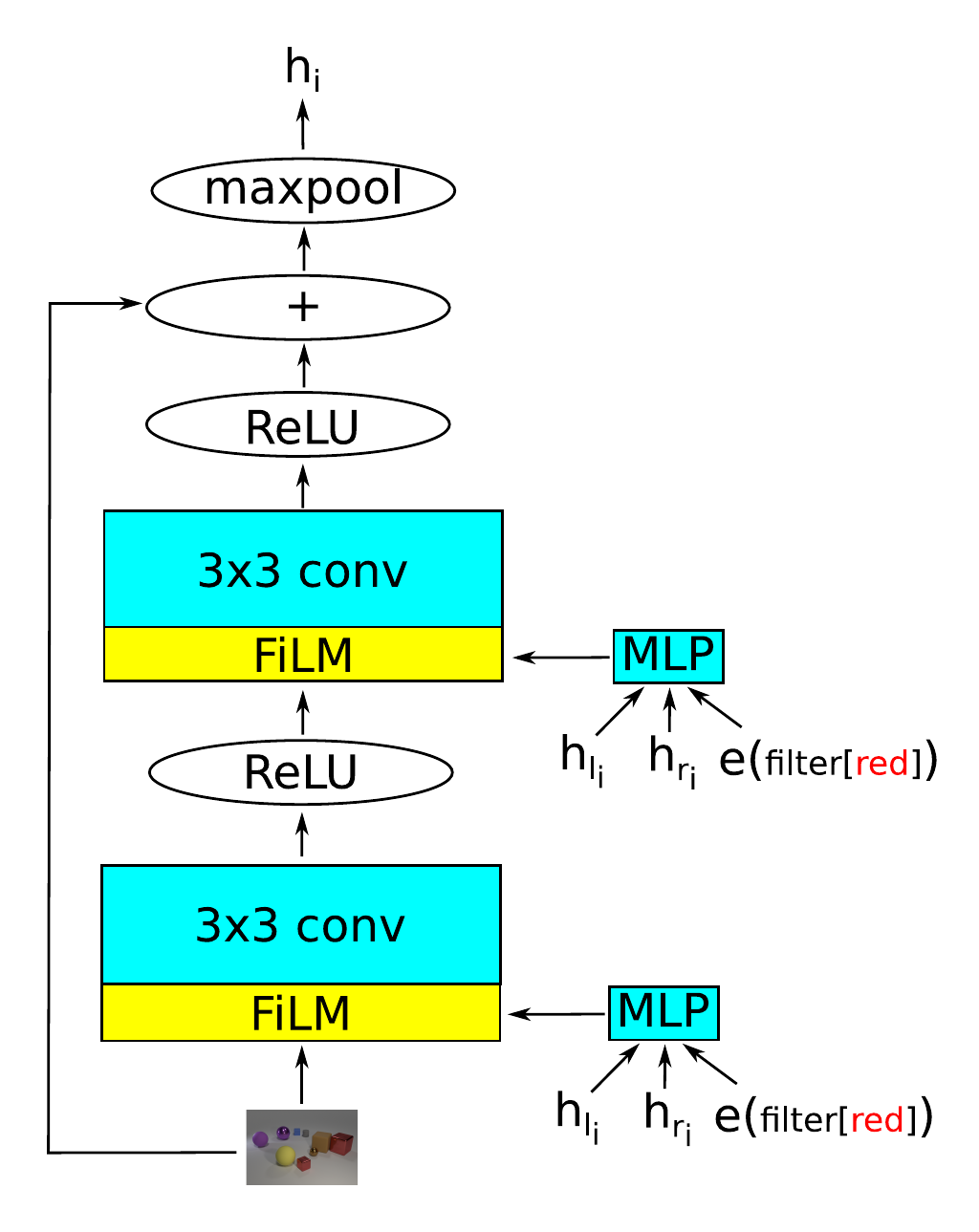} % cut the image to avoid extra page?
\caption{\label{fig:vectornmn}Computation performed by the Vector-NMN module. The 3D tensor $h_x$ represents the input image $x$. $h_{l_i}$ and $h_{r_i}$ are the outputs of the two preceeding modules $l_i$ and $r_i$ respectively. $e(p_i)$ is an embedding of the $i$-th program token. The weights of all matrix multiplications and convolutions (in cyan) are shared across all modules, whereas the unit-wise FiLM coefficients (in yellow) depend on the program token $p_i$ that the $i$-th module corresponds to.}
\end{figure}

See Figure \ref{fig:vectornmn} for a visual explanation of the Vector-NMN architecture. 

We conducted additional experiments to better understand why Vector-NMN outperforms Tensor-NMN. First, we considered a module architecture that differs from Tensor-NMN in that every module takes the image representation $h_x$ as an additional input. We called this model Tensor-NMN-Shortcut. Second, we evaluated a module architecture that is similar to Vector-NMN, but in which the module output $h_i$ is not max-pooled and remains a 3D tensor. To feed $h_i$ as an input to downstream modules, we concatenated it to the image representation $h_x$ in the same way as it is done in Tensor-NMN-Shortcut (note, that in Vector-NMN the pooled $h_i$ would instead be used to compute FiLM coefficients of the downstream module). We will refer to this model as Tensor-NMN-FiLM. These two additional module architectures, namely Tensor-NMN-Shortcut and Tensor-NMN-FiLM, allow us to test whether just connecting every module to the image representation (the former) or sharing weights across modules (the latter) would alone be sufficient to achieve generalization comparable to that of Vector-NMN. 

\begin{figure}
\centering
\includegraphics[width=0.5\textwidth]{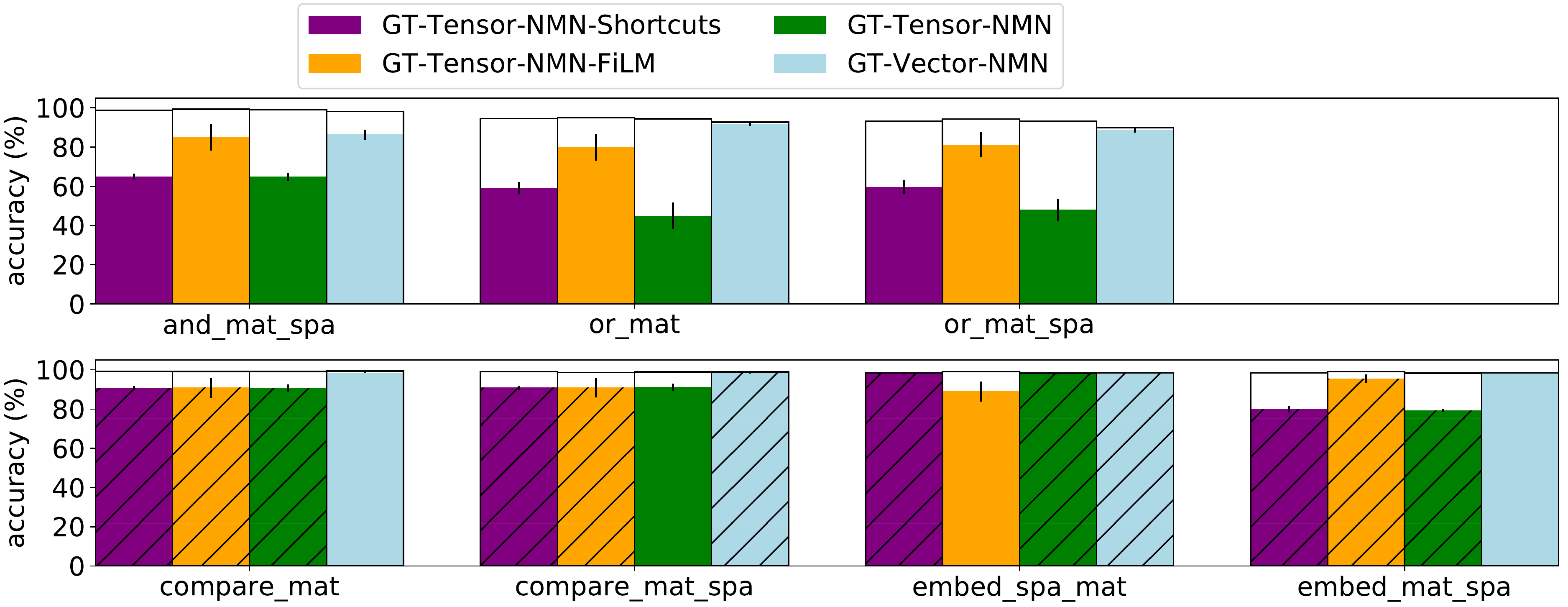}
\caption{\label{fig:ablation}Comparing different module architectures. See Appendix \ref{app:vectornmn} for more information.}
\end{figure}

The results are reported in Figure \ref{fig:ablation}. Both Tensor-NMN-Shortcut and Tensor-NMN-FiLM fall short of achieving their target performances on CLOSURE tests. This suggests that max-pooling the tensor-shaped module output and thereby turning it into a vector is indeed a key cause of the Vector-NMN's better generalization.

\end{document}